\documentclass[final,5p,times,twocolumn]{elsarticle}

\usepackage{geometry}
\usepackage{amsmath,amssymb}
\usepackage{booktabs}
\usepackage{epstopdf}
\usepackage{graphicx}
\usepackage[hidelinks]{hyperref}
\usepackage[ruled,linesnumbered]{algorithm2e}

\journal{Journal Name}

\biboptions{numbers,sort&compress}

\begin{document}
	
	\begin{frontmatter}
		
		\title{Toward General Semantic Chunking: A Discriminative Framework for Ultra-Long Documents}
		\author{Kaifeng Wu}
		\author{Junyan Wu}
		\author{Qiang Liu\corref{cor1}}
		\author{Jiarui Zhang}
		\author{Wen Xu}
		\cortext[cor1]{Corresponding author.}
		\address{KingSoft Office Zhuiguang AI Lab}
		
		\begin{abstract}
			Long-document topic segmentation plays an important role in information retrieval and document understanding, yet existing methods still show clear shortcomings in ultra-long text settings. Traditional discriminative models are constrained by fixed windows and cannot model document-level semantics; generative large language models can output paragraph boundaries, but inference is expensive and long inputs are difficult to support. To address these issues, we propose a discriminative segmentation model based on Qwen3-0.6B. On top of the backbone network, we add a cross-window context fusion layer and a boundary classification head, and combine them with an overlapping sliding-window strategy. Our model supports single-pass inputs of up to 13k tokens and can be extended to ultra-long documents for paragraph boundary detection. To further enhance downstream retrieval efficiency, we derive a vector fusion method with scalar correction, which compresses the representation of ultra-long segments into a single vector without semantic loss. Experiments on the Wikipedia long-document topic segmentation dataset WIKI-727K show that, compared with three generative models based on Qwen2-0.5B released by Jina, our method achieves a better macro-averaged F1 and delivers two orders of magnitude faster inference, substantially improving the practicality and scalability of long-document processing.
		\end{abstract}
		
		
	\end{frontmatter}
	
	\section{Introduction}
	Topic segmentation aims to divide a long text into several semantically continuous and topically coherent segments. It is a key step in document understanding and structural analysis, and has been widely used in downstream tasks such as information extraction, summarization, retrieval, and long-document reading assistance \cite{Shtekh2018,Xiao2019,Liu2022}. Unlike coarse segmentation based on paragraphs or fixed lengths, topic segmentation requires a model to automatically identify potential topic transition points. Therefore, it must jointly model local semantic changes, discourse structure, and global cross-segment coherence.

	Early work mainly relied on unsupervised methods, such as computing lexical or semantic similarities between adjacent text units or using topic models to infer potential boundaries \cite{Hearst1997,Riedl2012}. These approaches typically depend on hand-crafted heuristics, are sensitive to parameters and thresholds, and struggle to stably capture semantic shift signals in complex contexts. With the emergence of large-scale annotated data, topic segmentation has gradually been treated as a supervised learning problem \cite{Badjatiya2018}. A representative benchmark is WIKI-727K proposed by Koshorek et al.\ \cite{Koshorek2018}, which transforms Wikipedia articles' section structure into a sentence-level boundary labeling task, enabling models to learn topic transitions from large-scale training data. On this benchmark, supervised methods significantly outperform unsupervised approaches \cite{Badjatiya2018,Koshorek2018}.

	Existing supervised topic segmentation methods can be roughly categorized into two types: sentence-pair discriminative models and sequence labeling models. The former determines whether adjacent sentence pairs belong to the same topic \cite{Wang2017}, while the latter assigns a boundary label to each sentence in a document \cite{Koshorek2018}. With pretrained language models such as BERT, both types have achieved substantial improvements. However, traditional pretrained models (e.g., BERT) are constrained by fixed input length and the quadratic complexity of self-attention, making it difficult to efficiently process ultra-long documents. To mitigate this, prior work has explored hierarchical architectures or sliding-window strategies: for example, the SECTOR model aggregates sentence-level representations for segment classification \cite{Arnold2019}, cross-segment attention is introduced to enhance context modeling \cite{Lukasik2020}, and other work applies sliding-window truncation to long texts \cite{Lo2021}. Although these methods extend the processable length to some extent, they still face a trade-off between global context and local details; information loss at window boundaries can hurt segmentation accuracy \cite{Xing2020}.

	Text coherence is also widely considered an important cue for topic segmentation. Recent studies introduce auxiliary objectives beyond the main task, such as neighboring sentence order prediction, contrastive learning, or semantic consistency modeling, to enhance sensitivity to topic transitions \cite{Barrow2020}. For example, the CATS framework combines a two-level Transformer with coherence modeling, improving detection of topic changes via sentence order perturbations and contrastive learning \cite{Somasundaran2020}; and incorporating semantic similarity tasks over consecutive sentence pairs can strengthen consistent contextual representations \cite{Yu2023}. While these methods improve local discrimination, they remain constrained by input length and computational efficiency when dealing with ultra-long documents.
	
	Meanwhile, generative large language models (LLMs) provide a new perspective on topic segmentation. With carefully designed prompts, a model can be guided to ``generate'' segments or boundary labels, showing strong expressive capability on small-scale examples or short texts. However, this line of work still has clear limitations for topic segmentation: (1) high inference cost and slow speed---token-by-token generation is hard to satisfy large-scale or real-time needs; (2) limited robustness for long-text modeling---even though long-context modeling has improved significantly in recent years, performance still degrades notably when inputs exceed several thousand tokens, making it difficult to stably identify topic shifts and their boundaries; (3) unstable outputs that are highly sensitive to prompt design. Prior studies indicate that even with carefully crafted prompts, generative models such as ChatGPT underperform discriminative models fine-tuned on large-scale data (e.g., Longformer) for long-document topic segmentation~\cite{Devlin2019,QwenTechReport2025}. These observations suggest that relying solely on generative LLMs is not an ideal solution.
	
	Based on the above analysis, we propose a discriminative topic segmentation model for ultra-long documents. We use Qwen3 with 0.6B parameters as the backbone network, and introduce a lightweight Transformer encoder on top for cross-segment context fusion. An MLP classification head is used to directly predict paragraph boundary probabilities. With an efficient discriminative architecture, our model natively supports long sequences of more than 13k tokens, significantly breaking the length bottleneck of traditional pretrained models. For even longer documents, we further adopt an overlapping sliding-window strategy to mitigate information loss near window boundaries. In addition, since boundary samples are far fewer than non-boundary samples, we apply loss re-weighting during training to improve recall for true boundaries. Overall, our method leverages the semantic representation power of large models while retaining the efficiency and stability of discriminative modeling.
	
	Our main contributions are summarized as follows:
	\begin{itemize}
		\item We propose a discriminative topic segmentation framework supporting 13k+ input length, effectively breaking the length limitation while maintaining efficient inference.
		\item We design a context fusion layer and an overlapping sliding-window mechanism to capture local semantic changes while preserving global discourse information, and improve boundary prediction stability via class-imbalance optimization.
		\item We introduce a mathematically equivalent vector fusion strategy for ultra-long chunk storage, reducing the retrieval complexity from $O(N)$ to $O(1)$ while strictly preserving cosine similarity.
		\item We conduct systematic evaluations on WIKI-727K; results show that, compared with generative segmentation approaches, our method achieves substantial gains in F1 and recall, together with nearly two orders of magnitude inference speedup. Although precision decreases slightly, the overall segmentation performance becomes more balanced and robust.
	\end{itemize}
	
	\section{Method}
	\subsection{Model Architecture}
\begin{figure}[t]
		\centering
		\includegraphics[width=0.92\linewidth]{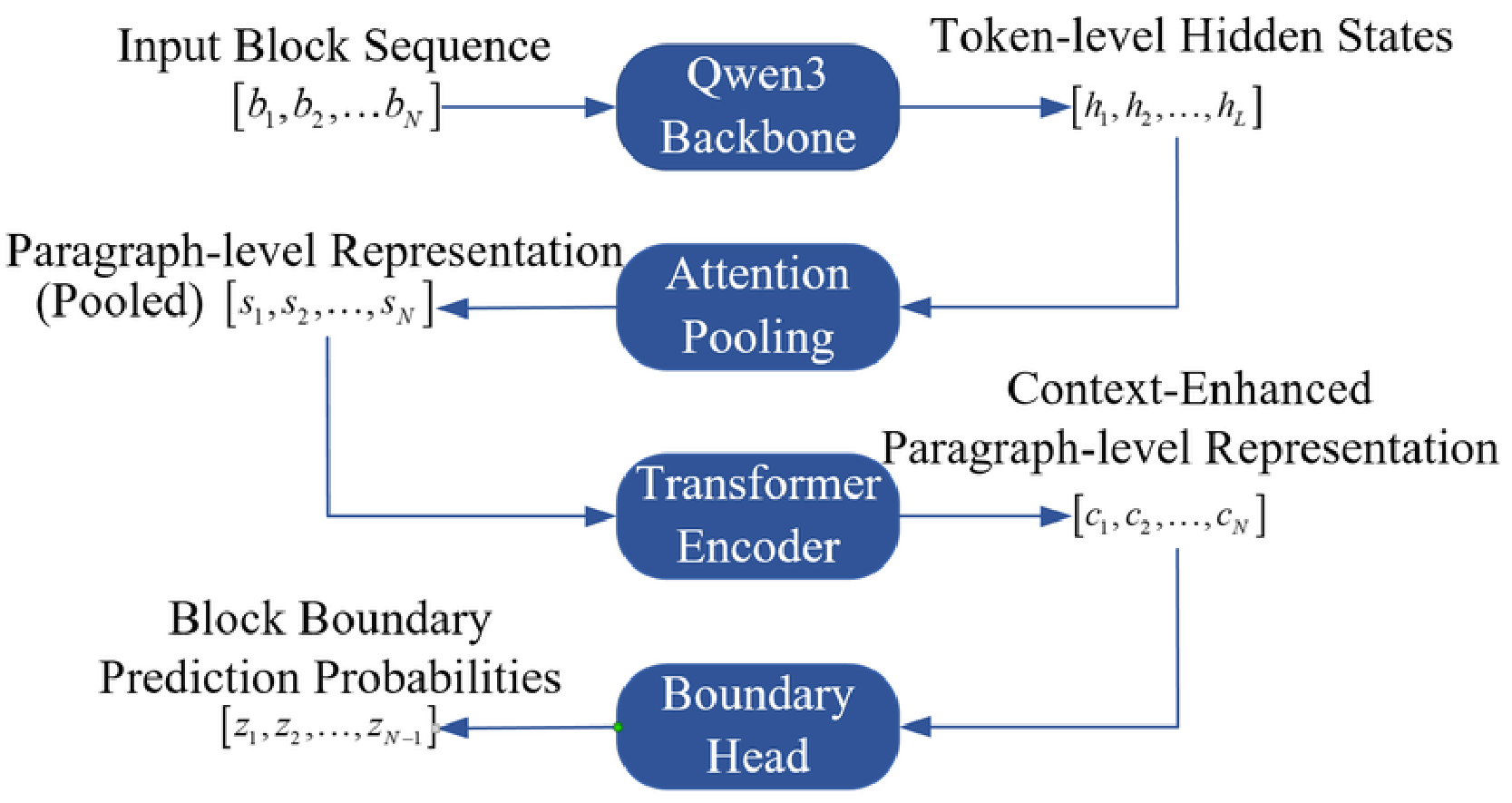}
		\caption{Overall architecture of the proposed discriminative topic segmentation model.}
		\label{fig:arch}
	\end{figure}

	The overall architecture of the proposed semantic topic segmentation model is illustrated in Fig.~\ref{fig:arch}. The model follows a hierarchical design with five stages: block input, token-level semantic encoding, block-level semantic aggregation, cross-block context modeling, and discriminative boundary prediction. This design enables accurate modeling of topic boundaries in long documents.
	
	Given an input document, we first split it into a sequence of contiguous text blocks (blocks)
	\[
	[b_1, b_2, \dots, b_N],
	\]
	where each block $b_i$ typically corresponds to a sentence or a paragraph, and the boundary between adjacent blocks is treated as a potential topic transition position. All blocks are concatenated in their original order and fed into the Qwen3 pretrained language model as the encoding backbone. After encoding, the model outputs token-level hidden states
	\[
	[h_1, h_2, \dots, h_L],
	\]
	where $L$ denotes the total number of tokens in the document.
	
	To obtain block-level semantic representations from token-level states, we introduce an \emph{attention pooling} module within each block. It aggregates token hidden states belonging to the same block with learned weights, producing a sequence of block representations
	\[
	[s_1, s_2, \dots, s_N].
	\]
	Attention pooling automatically focuses on tokens that are more critical to semantic expression, yielding compact yet discriminative block representations.
	
	We then feed the block representations into a Transformer encoder to model contextual dependencies between blocks. This context modeling module captures long-range cross-block semantic relations and outputs context-enhanced block representations
	\[
	[c_1, c_2, \dots, c_N].
	\]
	Through this process, the model leverages both local semantics and global document structure to support boundary decisions.
	
	Finally, a discriminative boundary prediction head (\emph{Boundary Head}) predicts whether a topic transition exists between adjacent blocks. For each adjacent pair $(c_i, c_{i+1})$, the head outputs a scalar probability, forming
	\[
	[z_1, z_2, \dots, z_{N-1}],
	\]
	where $z_i \in [0,1]$ denotes the predicted probability that a topic boundary exists between $b_i$ and $b_{i+1}$. Unlike generative or sequence-decoding-based approaches, our model directly models inter-block boundaries in a discriminative manner, avoiding complex decoding and substantially improving inference efficiency while maintaining segmentation accuracy.
	
	\subsection{Input Modeling and Ultra-Long Document Handling}
	The model takes the block sequence $[b_1, b_2, \dots, b_N]$ as the basic input unit; thus, the block construction directly determines the candidate boundary set. In this work, topic segmentation is formulated as a discriminative problem---judging whether a semantic shift occurs between adjacent blocks. Accordingly, the model predicts boundaries only between adjacent blocks $(b_i, b_{i+1})$, turning topic segmentation into a standard sequential discrimination task. Therefore, we first specify how basic blocks are constructed and then define candidate boundaries accordingly.
	
	\subsubsection{Basic Text Block Modeling}
	Following the annotation scheme of WIKI-727K and our task setting, we adopt sentence-level splitting as the basic block unit. Specifically, an input document is decomposed into a sentence sequence
	\[
	[b_1, b_2, \dots, b_N],
	\]
	where each $b_i$ is a complete sentence, and positions between adjacent sentences are treated as candidate topic boundaries. The model then predicts a segmentation probability $z_i$ for each candidate boundary.
	
	Sentence-level blocks have clear modeling advantages. First, a sentence is usually considered the smallest semantically complete unit; compared with paragraph-level splitting, it provides finer granularity and covers most true topic transition positions, thus offering a sufficiently rich and uniform candidate boundary space. Second, this splitting naturally aligns with the binary, sentence-level boundary annotations in WIKI-727K, forming a one-to-one correspondence between prediction positions and supervision signals, which facilitates stable training and efficient optimization. Finally, sentence-level splitting does not rely on manually chosen segment lengths or heuristics, reducing subjective bias introduced by varying document structures and improving generality across text types.
	
	Note that sentence-level splitting does not assume that topic transitions must occur at sentence boundaries. Rather, it provides a well-covered and structurally clear candidate set; the model learns to distinguish semantic continuation from semantic shift over this set, combining topic segmentation with pretrained language models' sequence representation ability.
	
	\subsubsection{Overlapping Sliding Window Modeling for Ultra-Long Documents}
	Although Qwen3-0.6B supports long-context modeling of about 13k tokens, a small number of documents in real data can still exceed the model's maximum input length. To handle such ultra-long documents, we introduce an \textbf{overlapping sliding window} strategy that preserves cross-block semantic continuity as much as possible within a limited context window.
	
	When the token length exceeds the model limit, we split the document into multiple subsequences whose length does not exceed the maximum, following the original order. Adjacent windows keep a fixed overlap ratio (about 10\% in our setting) to avoid unstable predictions when a potential boundary lies near a window edge with insufficient context. The model runs the full encoding and boundary prediction pipeline independently on each window, producing boundary probability sequences.
	
	To merge window-level predictions, we use a probability-averaging fusion strategy in the overlapped regions. A candidate boundary in an overlapped region may be predicted in multiple windows; we average its predicted probabilities across windows as the final value. For non-overlapped regions, we directly keep the original predictions. This strategy mitigates window boundary effects without additional parameters or complex post-processing, while maintaining high computational efficiency.
	
	During training, we apply the same sliding-window splitting strategy to ultra-long samples, and include the resulting subsequences in batches to keep training and inference consistent. Benefiting from Qwen3-0.6B's long-context capability, most documents can still be encoded as a whole without slicing. Compared with traditional BERT-based methods, this significantly reduces semantic information loss due to forced truncation~[1], providing more complete and coherent context for subsequent boundary modeling.
	
	\subsubsection{Heuristic Segmentation Strategy}
	After obtaining the model's predicted probabilities for topic boundaries between adjacent text blocks, it is still necessary to transform the continuous boundary discrimination results into final semantic chunks that satisfy practical application requirements. It is important to note that the training objective of the discriminative model focuses on boundary-level classification performance (such as F1 score), whereas downstream applications are more concerned with the overall usability of generated text chunks in terms of length constraints and semantic coherence. To bridge this gap, we introduce a set of parameter-free heuristic segmentation strategies based on the model outputs to perform structured decision-making and length constraint adjustments on the predicted boundaries.
	
	Overall, this strategy revolves around two goals: ``avoiding overly long chunks that affect usability'' and ``avoiding overly short chunks that disrupt semantic stability,'' and adjusts the segmentation results in three stages from coarse to fine.
	
	First, the text is initially segmented based on the boundary prediction probability threshold $T_1$. For adjacent text blocks $b_i$ and $b_{i+1}$, when the corresponding boundary prediction probability $z_i \ge T_1$, a topic boundary is inserted at this position, thereby obtaining a set of initial semantic chunks. This stage mainly relies on the model's ability to discriminate significant topic transition positions to generate coarse-grained and high-confidence segmentation results.
	
	On this basis, a recursive splitting mechanism for long blocks is introduced to constrain the maximum length of chunks. When the token count of a chunk exceeds a preset upper limit (e.g., 700), it is considered difficult to process directly in practical applications. At this time, the candidate boundary position with the maximum prediction probability within the chunk is searched, and segmentation is performed again at this position; the above process is repeated for the newly generated sub-blocks until the lengths of all chunks satisfy the upper limit constraint. This recursive strategy effectively controls the size of individual chunks while following the model's semantic judgment.
	
	Finally, an adaptive merging strategy for short blocks is introduced to avoid generating overly fragmented segmentation results. When the token count of a chunk is lower than a preset lower limit (e.g., 85), it is considered that its semantic content may not be sufficient to independently constitute a stable topic unit. At this time, the segmentation probabilities corresponding to the boundary positions adjacent to the left and right of the chunk are compared, and it is merged to the side with the smaller segmentation probability, that is, the side with stronger semantic continuity. In this way, the occurrence of isolated short blocks can be reduced without significantly destroying topic consistency.
	
\begin{figure}[t]
		\centering
		\includegraphics[width=0.92\linewidth]{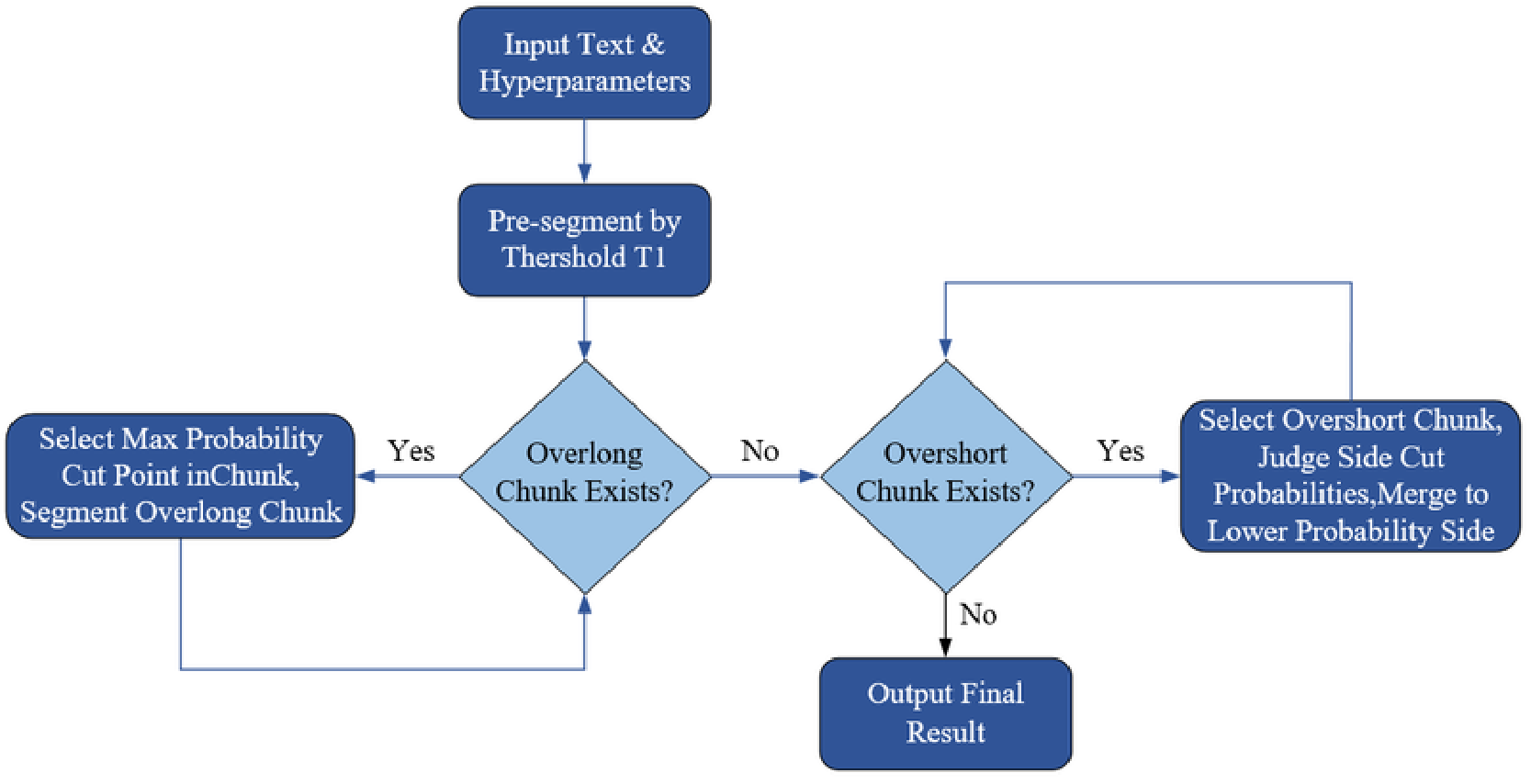}
		\caption{Overall process of the heuristic segmentation strategy.}
		\label{fig:heuristics}
	\end{figure}

	The above heuristic segmentation strategy combines discriminative boundary prediction results with practical engineering constraints without introducing additional training parameters or complex post-processing models, effectively improving the usability of final semantic chunks in terms of length distribution and structural stability. Its overall process is illustrated in Fig.~\ref{fig:heuristics}.
	
	\subsubsection{Vector Representation and Retrieval Adaptation for Ultra-Long Semantic Chunks}
	While the heuristic strategies described above ensure structural coherence, they may occasionally produce ``ultra-long chunks'' (e.g., resulting from the adaptive merging of multiple short blocks) that exceed the maximum context window of typical embedding models (e.g., 512 tokens for BERT-based retrievers). Truncating these chunks results in severe semantic information loss.
	
	A naive solution is to split such a long chunk $C$ into $N$ sub-segments $\{c_1, c_2, \dots, c_N\}$, encode them into vectors $\{v_1, v_2, \dots, v_N\}$, and compute the similarity with a query $Q$ by averaging the cosine similarities of all sub-segments:
	\begin{equation}
		\label{eq:avg_sim}
		\text{Sim}(C, Q) = \frac{1}{N} \sum_{i=1}^{N} \cos(v_i, v_q)
	\end{equation}
	where $v_q$ is the query vector. However, this approach increases storage costs linearly by $N$ and significantly increases retrieval latency, as $N$ dot products are required for a single document unit.
	
	To address this, we propose a \textbf{Vector Fusion with Scalar Correction} method. Our goal is to derive a single representative vector $V_f$ and a scalar correction factor $k$ such that the weighted cosine similarity of $V_f$ equals the average similarity of the sub-segments (Eq.~\ref{eq:avg_sim}).
	
	Let $\hat{v}_i = \frac{v_i}{\|v_i\|}$ and $\hat{v}_q = \frac{v_q}{\|v_q\|}$ be the normalized unit vectors. Equation~\ref{eq:avg_sim} can be rewritten as:
	\begin{equation}
		\text{Sim}(C, Q) = \frac{1}{N} \sum_{i=1}^{N} (\hat{v}_i \cdot \hat{v}_q) = \left( \frac{1}{N} \sum_{i=1}^{N} \hat{v}_i \right) \cdot \hat{v}_q
	\end{equation}
	Let $S = \sum_{i=1}^{N} \hat{v}_i$ be the sum of the normalized sub-vectors. We can express $S$ in terms of its magnitude and direction: $S = \|S\| \cdot \frac{S}{\|S\|}$. Substituting this back:
	\begin{equation}
		\text{Sim}(C, Q) = \frac{\|S\|}{N} \left( \frac{S}{\|S\|} \cdot \hat{v}_q \right) = \frac{\|S\|}{N} \cos(S, Q)
	\end{equation}
	Thus, we can define our stored representation as a tuple $(V_f, k)$, where:
	\begin{equation}
		V_f = \sum_{i=1}^{N} \frac{v_i}{\|v_i\|}, \quad k = \frac{\|V_f\|}{N}
	\end{equation}
	
	The final retrieval score is then computed as $\text{Score} = k \times \cos(V_f, Q)$. The detailed implementation is summarized in Algorithm~\ref{alg:vector_fusion}.
	
	\begin{algorithm}[ht]
		\caption{Vector Fusion with Scalar Correction (VFSC)}
		\label{alg:vector_fusion}
		\KwIn{Sub-segment vectors $\{v_1, \dots, v_N\}$ of an ultra-long chunk $C$.}
		\KwOut{Fused representative vector $V_f$ and scalar correction factor $k$.}
		
		\BlankLine
		\tcp{Offline: Encoding and Fusion}
		$S \leftarrow \mathbf{0}$\;
		\ForEach{$v_i$ \textbf{in} $\{v_1, \dots, v_N\}$}{
			$\hat{v}_i = v_i / \|v_i\|$ \tcp*{Unit vector conversion}
			$S = S + \hat{v}_i$ \tcp*{Directional accumulation}
		}
		$V_f = S$\; 
		$k = \|S\| / N$\;
		
		\BlankLine
		\tcp{Online: Efficient Retrieval}
		Given a query $Q$ with vector $v_q$\;
		$\text{Score} = k \times \text{cosine\_similarity}(V_f, v_q)$\;
		\Return \text{Score}\;
	\end{algorithm}
	
	This derivation proves that we can strictly reconstruct the average cosine similarity of an arbitrary number of sub-blocks using a single stored vector and one scalar. This method reduces storage complexity from $O(N \times d)$ to $O(d + 1)$ and retrieval computational complexity from $O(N)$ to $O(1)$, elegantly solving the ultra-long chunk storage problem without semantic degradation.
	
	\section{Experimental Setup}
	\subsection{Dataset}
	We use WIKI-727K as the main evaluation dataset~[2]. It is automatically constructed from Wikipedia articles and contains about 727,000 documents, with 582k/72k/73k documents in the training/validation/test splits, respectively. Each document provides paragraph boundary annotations derived from the original section headings, making it one of the most widely used benchmarks for topic segmentation~[21].
	
	Following common practice~[20], we formalize topic segmentation as a sentence-level sequence labeling problem. Given a document consisting of sentence sequence
	\[
	s_1, s_2, \ldots, s_n,
	\]
	the model predicts a binary label for each sentence $s_i$ (except the last one), indicating whether a topic boundary occurs after it. Label ``1'' means a topic shift happens after $s_i$, while label ``0'' indicates continuation of the current topic~[22].
	
	Since WIKI-727K covers open-domain topics and spans a wide range of document lengths, it provides a stable foundation for evaluating topic segmentation in long-document settings.
	
	\subsection{Baselines}
	To evaluate the effectiveness of our approach, we compare it with three small language models (SLMs) released by Jina based on Qwen2-0.5B. The three models are \texttt{simple-qwen-0.5}, \texttt{topic-qwen-0.5}, and \texttt{summary-qwen-0.5}, all designed for long-document automatic chunking and representative of the generative segmentation paradigm. The \texttt{simple-qwen-0.5} model uses basic structural cues to identify paragraph boundaries; \texttt{topic-qwen-0.5} performs segmentation via topic cues and inference, emphasizing semantic consistency; and \texttt{summary-qwen-0.5} predicts boundaries while generating segment summaries to support richer semantic expression.
	
	All three are trained under the same data condition and perform generative segmentation with a unified model structure, forming reasonable baselines for our discriminative model. We compare their results on the WIKI-727K test set to verify our performance advantages.
	
	\subsection{Implementation Details}
	The model is implemented in PyTorch, using Qwen3-0.6B as the pretrained backbone. During downstream training, the backbone is fine-tuned together with the newly added Transformer fusion layer and MLP classification head. Different learning rates are applied: \(6\times10^{-5}\) for the backbone, and \(5\times10^{-5}\) for the fusion layer and classification head. The model is fine-tuned for 16 epochs on the WIKI-727K training set with a global batch size of 128. Training adopts the AdamW optimizer with a weight decay of 0.01, and employs weighted cross-entropy loss with a positive-to-negative class weight ratio of 7:1 to mitigate class imbalance and enhance boundary recognition. Model selection is determined by F1 scores on the validation set, and F1 is also used as the primary evaluation metric, highlighting the model’s capability in boundary discrimination.
	
	\section{Results and Analysis}
	\subsection{Performance Comparison on WIKI-727K}
	We compare our method with multiple baseline models on the WIKI-727K test set; the overall results are shown in Table~\ref{tab:metrics}.
	
	For generative segmentation, among three prompt strategies based on Qwen2-0.5B, the chain-of-thought topic prompting strategy (\texttt{CoT Topic Chunking}) achieves the best overall performance with F1 of 0.5185, slightly higher than the summary prompting strategy (F1 = 0.5152), while the simple prompting strategy performs worse (F1 = 0.4702). The precision--recall distribution indicates distinct segmentation styles across prompts: the summary prompting method yields the highest precision (0.5668) but lower recall (0.5159), suggesting more conservative segmentation; in contrast, the chain-of-thought topic prompt reaches a more balanced trade-off between precision (0.5405) and recall (0.5388), leading to a better F1.
	
	Compared with generative methods, our Qwen3-0.6B-based discriminative model (with a custom neck) shows clear advantages. After supervised fine-tuning on WIKI-727K, it reaches precision of 0.4628 and recall of 0.7312, obtaining an F1 of 0.5503, improving the best generative baseline by about 3 percentage points. This suggests that the discriminative formulation substantially enhances boundary recall, which in turn drives a clear improvement in overall F1 performance.
	
	\begin{table}[t]
		\centering
		\caption{Evaluation metrics on the WIKI-727K test set.}
		\label{tab:metrics}
		\begin{tabular}{lccc}
			\toprule
			Model & Precision & Recall & F1-Score \\
			\midrule
			simple-qwen-0.5 & 0.5346 & 0.4496 & 0.4702 \\
			topic-qwen-0.5 & 0.5405 & 0.5388 & 0.5185 \\
			summary-qwen-0.5 & 0.5668 & 0.5159 & 0.5152 \\
			Qwen3-0.6B + custom neck & 0.4628 & 0.7312 & 0.5503 \\
			\bottomrule
		\end{tabular}
	\end{table}
	
	\subsection{Discriminative vs. Generative: Comparative Analysis}
	The results show that, on ultra-long text topic segmentation, discriminative models outperform generative methods in both overall performance and stability. This advantage mainly comes from end-to-end optimization on large-scale labeled data, which enables the model to directly learn conditions under which topic boundaries occur. In contrast, although generative LLMs have strong language understanding, the lack of explicit supervision for segmentation makes their boundary predictions often rely on salient surface topic changes, limiting their ability to detect fine-grained topic transitions.
	
	More specifically, generative methods exhibit a clear precision--recall trade-off under different prompt strategies: some prompts favor conservative segmentation and output boundaries only when topic shifts are obvious, leading to low recall; while prompts involving summarization or rewriting may produce overly uniform segmentation and introduce false positives, reducing precision. In contrast, the discriminative model, driven by explicit boundary supervision, is able to capture subtle topic transitions more effectively. After fine-tuning, it achieves a substantial improvement in recall, establishing a clear advantage on the recall dimension. Although this gain is accompanied by a slight drop in precision, the overall F1 score still increases, demonstrating that the model benefits from stronger boundary sensitivity without compromising overall segmentation quality.
	
	It is worth noting that even without fine-tuning on labeled data, the discriminative model shows much higher recall than generative methods, suggesting that pretrained large models already have some ability to identify potential topic changes. However, without precise supervision, this ability is hard to calibrate and may lead to many false positives. This further highlights the critical role of high-quality supervision data for behavior calibration in long-document segmentation tasks, consistent with prior findings~[17].
	
	Beyond performance advantages,, the discriminative model also offers clear advantages in inference efficiency and engineering usability. Leveraging Transformer parallel computation, it can process a $\sim$13k-token text within hundreds of milliseconds, with throughput comparable to BERT-like models. In contrast, generative models rely on token-by-token generation of boundaries or rewrites, making them slow and unsuitable for high-throughput applications. Moreover, the discriminative output form of our model facilitates seamless integration with existing information retrieval and summarization systems, reducing deployment cost and system coupling. Taken together, these properties indicate that the proposed approach holds strong practical potential for large-scale document processing scenarios.
	
	\section{Conclusion}
	We propose a Qwen3-0.6B-based discriminative Chunk model for ultra-long text topic segmentation. By introducing a cross-segment context fusion layer and a boundary classification head on top of a pretrained backbone, we model topic segmentation as binary classification between adjacent text blocks. The proposed method is structurally simple yet significantly improves long-document modeling capability and inference efficiency. It natively supports about 13k-token inputs and further mitigates window boundary effects in ultra-long documents via an overlapping sliding window and probability fusion strategy. Furthermore, we solve the length constraint of embedding models through a vector fusion mechanism, ensuring that generated ultra-long chunks can be indexed and retrieved efficiently. Experiments on WIKI-727K show that, compared with multiple generative chunking methods based on small language models, our model achieves a better macro-averaged F1, especially with clear advantages in recall and stability, while avoiding the high computational overhead of token-by-token decoding. Overall, we validate the effectiveness and engineering feasibility of discriminative modeling for ultra-long text topic segmentation, providing a practical solution that balances performance and efficiency for long-document structuring.
	
	\nocite{*}
\bibliography{ref}
\end{document}